\documentclass[sigconf]{acmart} 

\usepackage{balance} 

\usepackage{setspace}
\usepackage[multiple]{footmisc}
\usepackage{bigfoot}
\usepackage{amsmath}
\usepackage{array}
\usepackage{float}
\usepackage{algorithm}
\usepackage{mathtools} 
\usepackage{subcaption}
\usepackage{algpseudocode}
\usepackage{amsthm}
\algnewcommand{\LineComment}[1]{\State \(\triangleright\) #1}

\DeclareNewFootnote{AAffil}[arabic]
\DeclareNewFootnote{ANote}[fnsymbol]

\algdef{SE}[SUBALG]{Indent}{EndIndent}{}{\algorithmicend\ }%
\algtext*{Indent}
\algtext*{EndIndent}

\title{An Algorithm for Adversary Aware Decentralized Networked MARL}

\author{Soumajyoti Sarkar \footnoteAAffil{This work has been done while the author was at Arizona State University and \\prior to joining Amazon}}
\affiliation{
  \city{California}
  \country{United States}}
\email{sarkar.soumajyoti@gmail.com}

\begin{abstract}
Decentralized multi-agent reinforcement learning (MARL) algorithms have become popular in the literature since it allows heterogeneous agents to have their own reward functions as opposed to canonical multi-agent Markov Decision Process (MDP) settings which assume common reward functions over all agents. In this work, we follow the existing work on collaborative MARL where agents in a connected time varying network can exchange information among each other in order to reach a consensus. We introduce vulnerabilities in the consensus updates of existing decentralized MARL algorithms where some agents can deviate from their usual consensus  update, who we term as adversarial agents. We then proceed to provide an algorithm that allows non-adversarial agents to reach a consensus in the presence of adversaries under a constrained setting.
\end{abstract}

\newcommand{\BibTeX}{\rm B\kern-.05em{\sc i\kern-.025em b}\kern-.08em\TeX}

\begin{document}


\pagestyle{fancy}
\fancyhead{}


\maketitle 


\section{Introduction}

Multi-agent reinforcement learning constitutes a reinforcement learning scenario where multiple agents participate to jointly learn an optimal policy and where optimality is defined in terms of some objective  \cite{busoniu2008comprehensive}. In a collaborative setting, the agents have to work together with a goal to optimize a shared reward metric \cite{kok2006collaborative}. Multi-agent environments are inherently non-stationary since the other agents are free to change their behavior as they also learn and adapt and this makes designing algorithms for MARL more complex that single agent systems. Our model differs from the  traditional collaborative and fully co-operative multi-agent systems in that the agents in our system are heterogeneous and can have different local reward functions. We work on sequential decision-making problems in which the agents repeatedly interact with their environment and try to jointly optimize the long-term reward they receive from the system, which depends on a sequence of decisions made by them and the signals shared by other agents in the network.

Existing value-factorized based MARL approaches perform well in various multi-agent cooperative environments under the centralized training and decentralized execution (CTDE) scheme, where all agents in a team are trained together by the centralized value network and each agent executes its policy independently \cite{lyu2021contrasting, lowe2017multi}. One of the existing issues in such systems is that in the centralized training process, the environment for the team is partially observable and non-stationary. The observation and action information of all the agents cannot represent the global states. What follows is that the existing methods perform poorly and are highly sample inefficient. In these regimes, regret minimization is a promising approach as it performs well in partially observable and fully competitive settings. However, it tends to model others as opponents and thus cannot work well under the CTDE scheme. In our work, we set out to respect two constraints: first, we want the agents to be fully decentralized even in their training and second, we follow the model free actor critic type algorithms for our collaborative MARL setup as opposed to Q-learning type algorithms or centralized training with model based approaches \cite{willemsen2021mambpo}.

The major difference in the kind of algorithms we follow in this paper with that of CTDE techniques is that our work follows the studies on consensus based distributed optimization \cite{nedic2009distributed} where the agents exchange parameters instead of their actions and states with other agents \cite{lowe2017multi}. There is no centralized training and each agent performs the critic step independently. These consensus based systems have been popular in many fields like sensor networks, social learning, co-ordination of vehicles and co-ordinated synchronous distributed optimization to name a few. However, most of these studies evade the risk of adversaries or agents that deviate from consensus updates. For example, in studies where humans participate in decision making and social learning over time \cite{celis2017distributed, sarkar2020use}, presence of adversarial nodes could disrupt the learning dynamics. The study in \cite{figura2021resilient} showed that a single adversarial agent can persuade all the other agents in the network to implement policies that optimize an objective that it desires. It then becomes important to answer whether we can modify the existing consensus based decentralized MARL algorithms to respond to the presence of malicious agents who deviate from the consensus rule updates or whether the non-malicious agents can still converge to the optimal solution in the presence of adversaries.

We study the setup where the rewards or the data for the agents are not corrupted but instead where the adversarial agents do not follow the consensus updates to obtain a joint optimal policy. We consider the scenario where there could exist more than a single adversarial agent in our setup. Section 2 provides the technical preliminaries of our networked MARL setup and Section 3 discusses the adversarial settings and the adversary aware consensus MARL algorithm. 


\section{Decentralized Networked MARL}
Our work is heavily influenced by the research on decentralized multi-agent reinforcement learning done in \cite{zhang2018fully}. The authors develop a consensus based reinforcement learning algorithm in a fully decentralized and networked setting. We primarily extend their work in a networked setting where malicious agents are present under a constrained setup. Our networked multiagent MDP is constructed as follows: we have a state space $\mathcal{S}$ shared by all agents $\mathcal{N}$ in a network, such that $\mathcal{A}^i$, $i \in |\mathcal{N}|$ is the action of agent $i$. Let $\mathcal{A} = \prod_{i=1}^N \mathcal{A}^i$ be the joint action space of all agents. Following this, $\mathcal{R}^i: \mathcal{S} \times \mathcal{A} \rightarrow \mathbb{R}$ is the local reward function of agent $i$ and $P: \mathcal{S} \times \mathcal{A} \times \mathcal{S} \rightarrow [0,1]$ is the state transition probability of the MDP. We assume that the states are observable globally in order to ensure co-operation among agents, but the rewards are observed locally as a function of the neighbors of the agent. At time $t$, an agent $i$ chooses an action $a^i_t$ given state $s_t$, following its own local policy $\pi^i: \mathcal{S} \times \mathcal{A}^i \rightarrow [0, 1]$, where $\pi^i(s, a^i)$ represents the probability of choosing action $a^i$ at state $s$. In this setup, the joint policy of all agents is given by $\pi(s, a) = \prod_{i \in \mathcal{N}} \pi^i (s, a^i)$.

For an agent $i$, the local policy is parameterized by $\pi^i_\theta$ where $\theta^i \in \Theta^i$ are the parameters and the joint policy is given by $\Theta = \prod_{i=1}^N \Theta^i$. As is standard in the assumptions of actor-critic algorithms with function approximation, the policy function $\pi_{\theta}^i(s, a^i) > 0$ for any $\theta^i \in \Theta^i$. We assume that $\pi_{\theta^i}^i(s, a^i)$ is continuously differentiable with respect to the parameter $\theta^i$ over $\Theta^i$. For all $\theta \in \Theta$, the transition matrix of the Markov chain $\{s_t\}_{t \geq 0}$ induced by policy $\pi_{\theta}$, for any $s, s'$ $\in \mathcal{S}$ is given by 

\begin{equation}
    P^\theta(s' | s) = \sum_{a \in \mathcal{A}} \pi_\theta(s, a) \cdot P(s'|s, a)
\end{equation}
It comes along with the assumption that $\{s_t\}_{t \geq 0}$ is irreducible and aperiodic under any $\pi_\theta$, and the chain of the state-action pair $\{(s_t, a_t)\}$ has a stationary distribution $d_\theta(s) \cdot \pi_\theta(s, a)$ for all pairs. \\

\noindent \textbf{Objective of the MARL:} The collective objective of the agents is to collaborate and find a policy $\pi_\theta$ such that it maximizes the average global long term rewards while only utilizing information that is local to the agents in the network. In that context, let $r_{t+1}^i$ denote the reward received by agent $i$ at time $t$. Then the goal of all agents collectively is to optimize the following objectives:

\begin{equation}
\begin{split}
    max_\theta J(\theta) & = lim_T \frac{1}{T} \mathbb{E}\Big( \sum_{t=0}^{T-1} \frac{1}{N} \sum_{i \in \mathcal{N}} r_{t+1}^i\Big) \\
    & = \sum_{s \in \mathcal{S}, a \in \mathcal{A}} d_\theta(s) \pi_\theta(s, a)\cdot \overline{R}(s, a).
\end{split}
\end{equation}
where $\overline{R}(s, a)$ = $\frac{1}{N} \cdot \sum_{i \in \mathcal{N}} R^i(s, a)$ is the globally averaged reward function. Following this, we have $\overline{R}(s, a) = \mathbb{E}[\overline{r}_{t+1} | s_t = s, a_t = a]$ where $\overline{r}_t = \frac{1}{N} \sum_{i \in \mathcal{N}} t_t^i$. In that context of the symbols, the global state value and action value functions can be denoted by $\mathcal{Q}_\theta(s, a) = \sum_{t} \mathbb{E} [\overline{r}_{t+1} -  J(\theta) \ | \ s_0 = s, a_0=a, \pi_\theta]$ and $\mathcal{V}_\theta(s) = \sum_{a \in \mathcal{A}} \pi_\theta(s, a)\mathcal{Q}_\theta(s,a)$. \\

\noindent \textbf{Policy Gradient with MARL:} To  develop an algorithm for MARL, we would apply the policy gradient theorem as mentioned in Theorem 3.1 in \cite{zhang2018fully}. For any $\theta \in \Theta$, let $\pi_\theta$ be a policy and $J(\theta)$ denote the globally averaged return,  we define the local advantage function $A_\theta^i: \mathcal{S} \times \mathcal{A}$ $\rightarrow \mathbb{R}$ as $A_\theta^i(s, a) = \mathcal{Q}_\theta(s, a) - \tilde{V}_\theta^i(s, a^{-1})$, where $\mathcal{Q}_\theta$ and $A_\theta$ are the corresponding global action-value and advantage functions and $\tilde{V}^i_\theta(s, a^{-1})$ is the local value function. The gradient of the policy is given by 

\begin{equation}
\nabla_{\theta^i} J(\theta) = \mathbb{E}_{s \sim d_\theta, a \sim \pi_\theta} [\nabla_{\theta^i} log \pi_{\theta^i}^i(s, a^i) \cdot A_\theta^i(s,a)] 
\end{equation}

\section{Adversary Aware MARL Algorithm}
\subsection{Non Adversarial Setting}
We first present the existing work under a non-adversarial setting. Here, he actor-critic based consensus algorithm for the MARL setup is as follows: we consider an agent specific local version of the global action-value function $Q_\theta$ which we denote as $\mathcal{Q}(\omega^i)$ where we hide the state, action factors in the function which is implicit and $\omega \in \mathbb{R}^U$, where the dimension $U$ is significantly less than the joint state action space. In order to use the policy gradient theorem discussed in the previous section to improve an agents' policy, each agent shares its local parameters $\omega^i$ with its neighbors on the network in order to reach an estimate of $\mathcal{Q}_\theta$ that is consensual among all agents in the network. Such distributed consensus algorithms have been proposed earlier \cite{nedic2009distributed} that guarantee convergence of the local agent functions.

\begin{algorithm}[!t]
\caption{Adversary aware Networked actor-critic algorithm }
\label{alg:edge_infer}
\begin{algorithmic}[1]
\State 	\textbf{Input:} Initial values of the parameters $\mu0^i$, $\omega_0^i$, $\tilde{\omega}_0^i$, $\theta_0^i$, $\forall i \in \mathcal{N}$; the initial state $s_0$ of the MDP and stepsizes $\{\beta_{\omega, t}\}_{t \geq 0}$ and $\{\beta_{\theta, t}\}_{t \geq 0}, F, V(N)$  
\State \textbf{Repeat:}

    \Indent
	\For { \textbf{all} $i \in \mathcal{N}$ }
        \State Observe state $s_{t+1}$ and reward $r^{i}_{t+1}$. 
        \State Update $\mu_{i+1}^i \leftarrow (1 - \beta_{\omega, t})$ $\cdot$ $\mu_t^i$ + $\beta_{\omega, t}$ $\cdot$ $r_{t+1}^i$.
        \State Select and execute action $a_{t+1}^i$ $\sim$ $\pi_{\theta_t^i}^{i}(s_{t+1}, \cdot)$.
        
    \EndFor

    \State Observe joint actions $a_{t+1}$  = $(a_{t+1}^1, \ldots , a_{t+1}^N)$.
    \For {\textbf{all} $i \in \mathcal{N}$}
        \State Update $\delta_t^i \leftarrow r_{t+1}^i$ - $\mu_t^i$ + $\mathcal{Q}_{t+1}(\omega_t^i)$ - $\mathcal{Q}_t(\omega_t^i)$.
        \State \textbf{Critic Step}: $\tilde{\omega}_t^i$ $\leftarrow$ $\omega_t^i$ + $\beta_{\omega, t}$ $\cdot$ $\delta_t^i$ $\cdot$ $\nabla_\omega\mathcal{Q}_t(\omega_t^i)$.
        \State \parbox[t]{\dimexpr\textwidth-\leftmargin-\labelsep-\labelwidth}{%
        Update $\mathcal{A}_t^i$ $\leftarrow$ $\mathcal{Q}_t(\omega_t^i)$ - $\sum_{a^i} \in \mathcal{A}^i$ $\pi_{\theta_t^i}(s_t, a^i$ $\cdot$ $\mathcal{Q}(s_t, a^i, \\ a^{-1}; \omega_t^i)$\strut}.
        \State Update $\psi_t^i$ $\leftarrow$ $\nabla_{\theta}^i$ \ log \ $\pi_{\theta_t^i}^i(s_t, a_t^i)$.
        \State \textbf{Actor Step}: $\theta_{t+1}^i$ $\leftarrow$ $\theta_t^i$ + $\beta_{\theta, t}$ $\cdot$ $\mathcal{A}_t^i$ $\cdot$ $\psi_t^i$.
        \State \parbox[t]{\dimexpr\textwidth-\leftmargin-\labelsep-\labelwidth}{%
        Send $\omega_t^i$ to the neighbors $\{ j \in \mathcal{N}: (i, j) \in \mathcal{E}_t \}$ over the \\ communication network $\mathcal{G}_t$\strut}.
    \EndFor
    
    \For {\textbf{all} $i \in \mathcal{N}$}
        \State Gather neighbors $K_i$ from $V(N)$
        \State \parbox[t]{\dimexpr\textwidth-\leftmargin-\labelsep-\labelwidth}{%
        Gather $\tau_i(t)$ by removing the highest and lowest \\ $F$ states among $K_i$\strut} 
        \State \textbf{Consensus Step}: $\omega_{t+1}^i$ $\leftarrow$ $\sum_{j \in \tau_i} c_t(i, j)$ $\cdot$ $\tilde{\omega}_t^j$.
    \EndFor
    \State Update the iteration counter $t$ $\leftarrow$ $t+1$.
    \EndIndent
\State \textbf{Until Convergence}
\end{algorithmic}
\end{algorithm}







\subsection{Adversary Aware Decentralized MARL}
In order to introduce the setup behind the MARL environment where malicious agents are a part of the network, we first define certain notations based on the setup of consensus based adversarial attacks in \cite{sundaram2015consensus}. We consider an undirected network $\mathcal{N} = {V(N), E(N)}$, where $V(N) = \{v1_, 1\ldots, v_n\}$ and $E(N)$ denotes the edges connecting pairs of nodes. Note that $v_i$ is just an exaggerated notation to denote an agent in a graph as opposed to $i$ that we have used in the previous sections but they refer to the same agent. Denoting $\mathcal{K}_i$ as the neighborhood vertices of $i \in V(N)$, for any $r \in \mathbb{N}$, a subset $S \subset V(N)$ is said to be \textit{r-local}, if $|\mathcal{K}_i \cap S| \leq r$ $\forall v_i \in V(N) \setminus S$. That is to say, for \textit{r-local} subset, there are at most $r$ nodes in the neighborhood of any vertex from $V(N) \setminus S$. For the setup of adversarial attacks in the multi-agent system, we consider the following: we consider that there are randomly chosen adversarial nodes in the network such that for each node $v_i$, there cannot be more than $g$ fraction of its neighbors who are adversaries, where $g \in [0, 1)$. This fraction $g$ is known to all nodes in the network, however the regular nodes do not know which or if any of their neighbor nodes are adversaries. Here we assume that the adversaries are restricted to form an \textit{F-local} set , where $F$ is a non-negative integer.

\subsection{Algorithm}
The overall actor-critic algorithm is detailed in Algorithm~\ref{alg:edge_infer}. In the actor step, each agent, each agent improves it policy as shown in Line 14, where $\beta_{\theta, t} > 0$ is the stepsize. Note that both the actor and the critic steps can be executed in a decentralized fashion without any centralized training. For the consensus step, one important thing to note that since the agents aim to optimize the globally averaged reward function $\overline{r}$, the agents share their local parameters $\omega^i$ with their neighbors and this allows the agents to improve their current policy using the policy gradient theorem. The parameter sharing involves a consensus update using the weight matrix $C_t = [c_t(i, j)]_{|\mathcal{N}| \times |\mathcal{N}|}$ such that $c_t(i, j)$ is the weight on the message sent by agent $j$ to agent $i$ at time $t$. An important restriction in our model is that we only consider $c_t(i, j) > 0$ if agent $i$ and $j$ are neighbors of each other in the network. Some discussions on the choice of $c_t(i, j)$ is in the later part of this section.

We modify Algorithm 1 in \cite{zhang2018fully} to incorporate co-ordinated responses by regular nodes in the presence of adversaries. At each time step, the regular nodes $\{v_i\}$ gather $\tilde{\omega}_t^j$ for all $v_j \in \mathcal{K}_i$ and remove the highest and lowest $F$ states and the remaining nodes are denoted by $\tau_i(t) \in \mathcal{K}_i$. The consensus action in the actor is then to  aggregate the parameters from the neighbors in $\tau_i(t)$. Note that, these updates are only done by  the regular nodes and the adversarial nodes aggregate the weights from the neighbors in any way they wish. The dynamics of updates by the regular nodes are local and decentralized since they do not require regular nodes to know anything beyond the signals sent from their neighbors. As mentioned in \cite{sundaram2015consensus}, when the network $\mathcal{N}$ is time-invariant, the effective neighbor set $\tau_i(t)$ is only a function of the states of neighbors of $v_i$ at time step $t$. This filtering is also closely connected to bandit based ranking of arms where the distribution of the means of the arms decide which arm would be picked, albeit here instead of picking one arm, the agent selects one or multiple arms \cite{even2006action}. The rest of the consensus style algorithm in this decentralized environment is the same as Algorithm 1 in \cite{zhang2018fully}.

The standard assumptions for the consensus matrix is defined in Assumption 4.4 of \cite{zhang2018fully} and while there can be many ways to define the weight, one popular way is to consider the notion of Metropolis weights \cite{xiao2005scheme}
\begin{align*}
    c_t(i, j) &= \Big\{ 1 + max[k_t(i), k_t(j)] \Big\}^{-1}, \ \ \forall (i, j) \in \mathcal{E}_t \\
    c_t(i, j) &= 1 - \sum_{j \in \tau_i} c_t(i, j) \forall i \in \mathcal{N}
\end{align*}
where $\mathcal{E}_t$ denotes the time varying edges and $k_t(i)$ is the degree of the agent $i$ in the time varying network.

\section{Related Work}
One of the assumptions in the process of social decision making is that individuals following a  learning trajectory (despite each individual having a limited memory) successfully converges to the best option for the collective population. While individuals participate in decision making where they associate different risk and rewards in an uncertain environment, they also tend to incorporate beliefs from their immediate neighbors in a networked environment, a phenomenon that has played a critical role in co-operative multi-agent systems \cite{celis2017distributed}. This begets the question as to how resilient are these consensus algorithms and how can agents adapt to decentralized training in the presence of malicious agents or adversaries. We briefly highly some notable studies done previously at the intersection of MARL and  \\

\noindent \textbf{MARL}: The field of MARL has evolved very rapidly over the past few years. The collective goal of the multi-agent system is to either reach a stable and consensus state for all agents \cite{fax2004information},
or solve a static optimization problem in a distributed fashion\cite{nedic2009distributed}. However, competition and collaboration always emerge between autonomous agents that learn by reinforcement over finite horizons. Some of the common approaches on modeling and solving cooperative multi-agent reinforcement learning problems include: (1) independent learners \cite{foerster2017stabilising, laurent2011world}, (2) fully observable critic \cite{mao2018modelling}, (3) value function factorization \cite{son2019qtran}, (4) consensus based RL \cite{cassano2020multiagent}, and (5) learning to communicate \cite{mordatch2018emergence}. \\

\noindent \textbf{Decentralized Networked MARL}: The challenge with fully co-operative multi-agent systems is that fully cooperative systems (Dec-POMDPs) are significantly harder to solve than single agent RL problems due to the combinatorial explosion in the joint action state spaces combining all agents. For this reason CTDE have become popular. The prevailing MARL paradigm of centralised training with decentralised execution (CTDE) [25,29,21] assumes a training stage during which the learning algorithm can access data from all agents to learn decentralised (locally-executable) agent policies. CTDE algorithms such as COMA \cite{foerster2018counterfactual} learn powerful critic networks conditioned on joint observations and actions of all agents. Other extensions of MADDP include shared experience actor-critic frameworks \cite{christianos2020shared} for efficient MARL exploration by sharing experience amongst agents as opposed to MADDPG which only reinforces an agent's own explored actions. There have also been studies conducted on making these RL systems in networked systems scalabale \cite{qu2020scalable}.\\

\noindent \textbf{Decision making for social systems}: There are two broad avenues of research on decision making in interactive environments considering social systems. The first group of studies focus on game theoretic environments where agents have similar adaptation and learning abilities and so the actions of each agent affect the task achievement of the other agents \cite{bowling2000analysis,nakayama2017nash,reverdy2014modeling}. The payoff of the agents in these games depend on whether they are purely collaborative or competitive or a mix. In a recent study on combining Reinforcement Learning with Agent Based Modeling \cite{sert2020segregation}, the authors address the self‑organizing dynamics of social segregation and explore the space of possibilities that emerge from considering different types of rewards. The second group of studies focuses on social influence from other agents as the intrinsic factor for the decision making. Social influence has been known to be an intrinsic factor in agents' choices \cite{sarkar2020use, schlag1998imitate}, and in recent studies, there have been attempts to propose a unified mechanism for coordination in MARL by rewarding  agents  for  having  causal  influence over other agents’ actions \cite{jaques2019social,xie2021learning}. Adversarial attacks during training can potentially impact the choices made by agents and impact the consensus algorithms as studied in \cite{figura2021resilient} and this is important to solve since attacks through social influence are a common tool for adversaries \cite{sarkar2019can}.

\section{Conclusion}
In this work, we discussed a simple strategy for agents to mitigate the nuances of adversarial agents in fully decentralized MARL where the agents are connected via a time varying network. We relied on the algorithm of \cite{zhang2018fully} and modified the  actor-critic algorithm in the presence of one or more adversaries. We did not provide a formal argument for the convergence of the consensus updates with our adversarial framework which we leave as one of the immediate next steps of this work. A second direction where this kind of work can be adapted is to use attention based actor critic algorithms \cite{iqbal2019actor} that modify the action value function by parameterizing it with agent specific attention weights. We can then use reward values to optimize for the attention weights over the time varying network.





\bibliographystyle{ACM-Reference-Format} 
\bibliography{sample}


\end{document}